\documentclass[10pt,letterpaper]{article}

\usepackage{iccm}
\usepackage{xurl}
\usepackage{amsmath,amssymb,amsfonts}
\usepackage{graphicx}
\iccmfinalcopy 

\usepackage{pslatex}
\usepackage{apacite}
\usepackage{float} 
\usepackage{nameref}

\title{Adapting A Vector-Symbolic Memory for Lisp ACT-R}
 
\author{{\large \bf Meera Ray (mfr5832@psu.edu)} \\
  Department of Computer Science and Engineering, The Pennsylvania State University \\
  W209 Westgate Building, University Park, PA 16802. USA
  \AND {\large \bf Christopher L. Dancy (cdancy@psu.edu)} \\
  Department of Industrial and Manufacturing Engineering, The Pennsylvania State University \\
  310 Leonhard Building, University Park, PA 16802. USA}

\begin{document}

\maketitle

\begin{abstract}
Holographic Declarative Memory (HDM) is a vector-symbolic alternative to ACT-R's Declarative Memory (DM) system that can bring advantages such as scalability and architecturally defined similarity between DM chunks. We adapted HDM to work with the most comprehensive and widely-used implementation of ACT-R (Lisp ACT-R) so extant ACT-R models designed with DM can be run with HDM without major changes. With this adaptation of HDM, we have developed vector-based versions of common ACT-R functions, set up a text processing pipeline to add the contents of large documents to ACT-R memory, and most significantly created a useful and novel mechanism to retrieve an entire chunk of memory based on a request using only vector representations of tokens. Preliminary results indicate that we can maintain vector-symbolic advantages of HDM (e.g., chunk recall without storing the actual chunk and other advantages with scaling) while also extending it so that previous ACT-R models may work with the system with little (or potentially no) modifications within the actual procedural and declarative memory portions of a model. As a part of iterative improvement of this newly translated holographic declarative memory module, we will continue to explore better time-context representations for vectors to improve the module's ability to reconstruct chunks during recall. To more fully test this translated HDM module, we also plan to develop decision-making models that use instance-based learning (IBL) theory, which is a useful application of HDM given the advantages of the system.

\end{abstract}

\section{Introduction}

In this paper, we describe our ongoing efforts to adapt HDM to work better with Lisp-based ACT-R, thus bridging a newer vector-symbolic model with decades of cognitive modeling done with Lisp ACT-R. Our eventual goal is to build a cognitive agent that simulates the actions taken by disaster survivors and the social factors that impact those decisions. We chose ACT-R due to the need for the cognitive agent to represent the cognitive processes of a disaster survivor well enough to accurately model the decisions they would make, relative to the empirical results gathered from human subjects; ACT-R has successfully modeled the decision making of humans in constrained experiments \cite{ritter_act-r_2019}, sometimes serving as a useful lower-level cognitive representation for rational level (i.e., \citeNP{newell_utc_1990}) decision-making theories (e.g., \citeNP{gonzalez_ibl_2003}). At the same time, the model needs to include situational awareness to accurately model how different survivors will respond to the same disaster environment differently depending on their social background \cite{prather_what_2022}. Here we use situational awareness to mean background knowledge, context, and worldview that puts the perceived environment into context \cite{lukosch_scientific_2018}. We hypothesize that a vector-symbolic memory architecture, Holographic Declarative Memory \cite{kelly_holographic_2020}, can be useful for representing this context  by reading in related texts into memory, which in turn influences which chunks are retrieved at runtime and which procedures are run. Furthermore, the increased scalability of a system like HDM may be more appropriate for representing the impact of existing social structures, which may themselves be more latently represented in memory in a way that requires a large number of concepts and associations to adequately represent.

 In the next sections, we first justify why it is worthwhile to use HDM when ACT-R already has a built-in Declarative Memory system. Next, we explain adaptations needed at the architectural level for (canonical, Lisp) ACT-R and HDM for integration. We conclude with a discussion of preliminary results of testing this new translated HDM and future work.

\subsection{What Does HDM add to ACT-R?}

Traditional ACT-R models have two main memory components for modelers to manipulate: procedural memory, typically production rules written by the developer, and declarative memory, which may store semantic or episodic information that is often entered by hand before the model runs \cite{ritter_act-r_2019}. ACT-R retrieves relevant memories based on environmental cues specified as requests in production rules. However, by default ACT-R's declarative memory (DM) system lacks partial memory recall, meaning a gradation of which chunks are relevant to a query, as opposed to returning either one matching chunk or none at all. Additionally, the time complexity to add and retrieve chunks as well as the space complexity to store chunks would be prohibitive to adding corpus-sized amounts of text to DM \cite{kelly_holographic_2020}. Holographic declarative memory (HDM) addresses these problems by representing each unique input token in vector space rather than storing the entire input, which allows for a continuous similarity measure between queries and memories. HDM can thus ``read" large corpora of relevant text into a model's memory; it is thus a distributed semantics model like Word2vec made available as a memory system for a cognitive architecture. Despite the machine-learning-like principle behind it, HDM still achieves high-level cognitive plausibility in free recall effects and significantly fits human performance in a decision task.

Thus, HDM presents a useful alternative to the default ACT-R declarative memory representation, especially for models that may need to represent sociocultural structures in memory and represent similarity between concepts or chunks in an architecturally-defined way. The latter ability is helpful for implementing theories in ACT-R that rely on context-dependent associations and similarities, such as Instance-Based Learning (IBL) Theory \cite{gonzalez_ibl_2003}. This vector-based representation may also make architecture-level integration with generative models (which typically also use vector-based representations for text-based ``tokens") more straightforward (e.g., see \citeNP{dancy_cogarch-genai_2023}).

Fortunately, the implementations of both HDM and the newest (Lisp) ACT-R provide enough of a foundation for a connection between the two systems. Nonetheless, to integrate HDM into the canonical version of ACT-R, both systems need to be adapted. This need for adaptation and expansion of HDM is especially true if one wants to use HDM while keeping declarative memory use patterns that are common amongst ACT-R modelers. 

\section{Adapting ACT-R for HDM}
We used the interface provided in the canonical ACT-R 7 Python connection files \cite{act-r_research_group_act-r_2023} and the HDM code Dr. Kelly wrote for Python-ACTR \cite{kelly_ecphoryhdm_2020} to implement the Lisp-side ACT-R commands with Python code. At present, we have implemented the core commands for adding to and retrieving from memory as well as convenience and utility commands like \verb|(sgp)| and \verb|(dm)|. The command \verb|(dm)| now prints the unique values stored by HDM rather than chunks, which are not explicitly stored, as well as a two-dimensional visualization of the HDM vector space. Goal chunks, which require more precise recall and typically do not come in a large scale number of chunks, are to be created with \verb|(define-chunk)| rather than \verb|(add-dm)| so ACT-R's default systems are used rather than HDM.  The parameters for the whole-chunk recall mechanism, which will be explored in the next section, can be set globally like any other parameter in \verb|(sgp)| or locally in any memory retrieval request.  

\subsection{Adding Text to HDM in ACT-R}
We added the ACT-R command \verb|(preprocess-text)|, which takes a raw plain text file and outputs a file ready to be added to memory. The preprocessing step uses the Natural Language Toolkit  (NLTK) to remove \textit{stopwords}, often-appearing words such as ``and" or ``the," which don't convey information specific to a task  \cite{sarica_stopwords_2021}. 

We made the decision to read the text into HDM  sentence by sentence, though \cite{kelly_holographic_2020} don't indicate whether or not this is necessary. When HDM encodes a list of values, it stores associations between all pairs of words in left-to-right order. For words that are far apart from each other, it is less worth computational resources to encode their associations with each other. Other ways of splitting the text may be optimal, but we chose the sentence as the unit to add at a time. Therefore, the preprocessing step tokenizes the text into sentences using NLTK's Punkt pretrained tokenizer to find sentence boundaries. 

\begin{figure}
    \centering
    \includegraphics[width=1\linewidth]{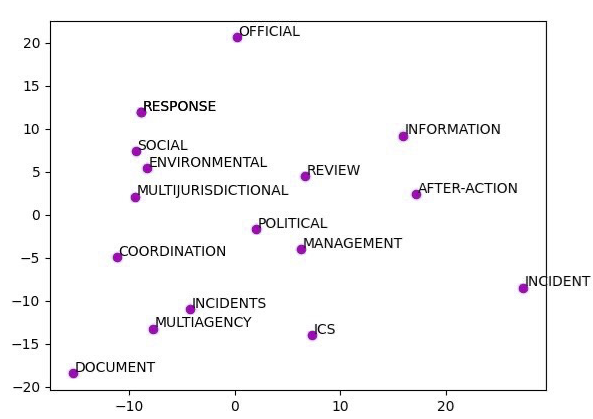}
    \caption{PCA Plot of HDM Vectors}
    \label{fig:nims-pca}
\end{figure}

From the preprocessed file, the modeler uses the \verb|(read-corpus-hdm)| command to read in the preprocessed file to HDM. As implemented, this step can only take place after the model has been loaded. An example of a plot of HDM vectors after being added to memory is Figure \ref{fig:nims-pca}. The corpus here is from a document instructing U.S. officials on how disaster response and recovery are organized.

\section{Adapting HDM for ACT-R} \label{adapting-hdm}

A downside of HDM is that, unlike ACT-R DM, it lacks a mechanism to retrieve a chunk, a group of memories bound together, in its entirety without knowing the slot corresponding to each unknown value being requested. For example, a chunk ``homeowner:yes damage:severe renter:no neighborhood:Eastville" could represent the status of an disaster suvivor agent's home after a flood; note the format of slot:value pairs. A cue to retrieve the chunk in DM would simply include a unique slot:value pair of the chunk — ``homeowner:yes". With HDM, one would have to specify that pair and then ask for one unknown at a time — ``homeowner:yes damage:?", ``homeowner:yes renter:?" and so on. We extended the HDM implementation using a simple chaining method so multiple unknowns can be requested, but every slot in the chunk still needs to be specified; this is cumbersome for a modeller compared to ACT-R's original DM. For us, it would require us to write more and longer production rules for our cognitive agent. Thus, we devised a method of full-chunk retrieval without storing chunks outright or deviating from HDM's memory vector approach. 
\newcommand{\cue}{\mathbf{q}}
\newcommand{\chunk}{\mathbf{c}}

We define the problem as follows: after adding chunk $\chunk = \{s_1{:}v_1, s_2{:}v_2, ... s_n{:}v_n\}$ into memory, retrieve the entire chunk with a cue $\cue = \{s'_1{:}v'_1, s'_2{:}v'_2, ... s'_m{:}v'_m\}$, where $\cue \subseteq \chunk$ and the slots and their associated values are $s_1 .. s_n$ and $v_1...v_n$ respectively. 

\subsection{Time Encodings}
\begin{figure}
    \centering
    \includegraphics[width=0.9\linewidth]{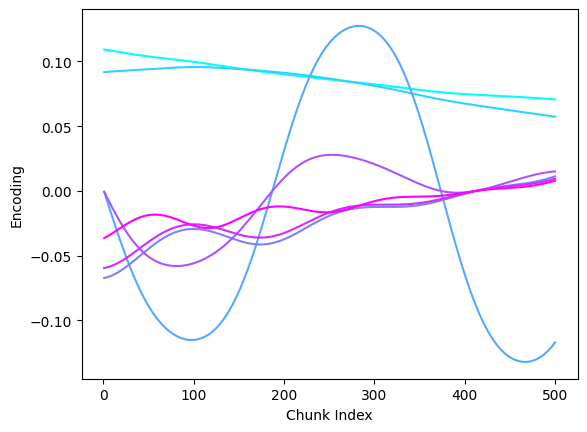}
    \caption{Oscillator function vector $\mathbf{T}(t)$ as a function of chunk index $t$. The functions plotted above are $T_1(t), T_{51}(t), T_{101}(t), T_{151}(t), T_{201}(t), T_{251}(t),T_{301}(t)$ out of the $\mathbf{T}(t)= [T_1(t), T_2(t),...T_{320}(t)] $. Note how the oscillators vary from lower to higher frequencies. While some individual functions repeat in encoding value over the chunk index range, the vector at any one chunk index is unique.}
    \label{fig:oscillators}
\end{figure}
First, we needed a way to represent which bits of memory were entered together as a chunk at the same time. We turned to neural oscillator encodings of serial memory \cite{brown_oscillator-based_2000} for a biologically-plausible theory of temporal encoding. In their model, there are fifteen oscillator functions $O_1(t)...O_{15}(t)$ of timestep $t$ which range from lower to higher frequency, with $R$ and $\phi$ as sources of noise:

\begin{align} 
    O_i &= sin(\phi + t\theta_i) \label{browntimevecs1}\\
    \theta_i &= S R 2^i \label{brown-theta}\\
    R &\sim \mathcal{N}(0, \sigma^2) \label{brown-normal}\\
    \phi &\sim \mathcal{U}(0, \pi / \theta) \label{browntimevecs4}
\end{align}

$S$ is a scaling parameter that can be swapped out for different time scales. \citeA{brown_oscillator-based_2000} sets $S = 10^{-5}$, but we've found different values  work better on different time scales (e.g. number of chunks). For the variance $\sigma^2$ of $R$, $\sigma^2 = 1$ in the Brown model but can be increased or decreased to increase or decrease the amount of noise.

The purpose of having noisy representations rather than simply, for example, sequentially numbering each chunk index, is to model the recall of human subjects in serial order memory tasks. An incorrect item is more likely to be exchanged for a correct item during recall the closer the two items are in position. When positions are hierarchical, e.g. lists of lists,  often the position of an item within a list is recalled correctly even if the list's position compared to other lists is not. For example, one may recall an item being in the third value of the second chunk instead of the third value in the first chunk. Multiple oscillating functions  represent this hierarchical error by creating ``bumps" in similarity between cognitive representations of two items even as their similarities decrease overall as their positions grow further apart.

To work more simply within the HDM framework, we flattened \citeA{brown_oscillator-based_2000}'s sixteen ``learning context" vectors of twenty elements each into a single 320-element time vector $\mathbf{T}$ for each timestep. Each element $T_i$ of the time vector in Equation \ref{oscillator-prod} randomly draws four of the 15 oscillators $O_j$ from Equation \ref{browntimevecs1}, with cosine being randomly substituted for sine with a probability of half:
\begin{align}
    T_i = \prod_{j=1}^{4} \sin(O_j) \label{oscillator-prod}
\end{align}

The function $\mathbf{T}(t)= [T_1(t)...T_{320}(t)]$ is generated and stored when the model is loaded. For each chunk added to memory, the time step $t$ increases by 1 and the time vector  $\mathbf{T}(t)$ is calculated. Note that the four oscillator functions randomly chosen above are not chosen with uniform probability; they're selected so the low-frequency oscillators make up a larger share than the higher-frequency oscillators. We did this because we found too many high-frequency oscillators led to nearly-repeating vectors over a small period of time. The probability distribution we selected was a discretized and bounded modification to the exponential distribution. We found it to be the best match for how the oscillators are distributed in Figure 6 of \citeA{brown_oscillator-based_2000}. The best match for the figure is with a scaling parameter of $\beta = 5.125$. However, the distribution can be used with different scaling parameters for learning context vectors of arbitrarily chosen sizes if one adjusts the scaling parameter to be the expected value of which oscillators is chosen. For example, the scaling parameter $\beta$ being $5.125$ means the center of the probability mass is between the sixth and seventh oscillator (zero indexing). A plot of an oscillator vector is shown in Figure \ref{fig:oscillators}. We used the default parameters described earlier ($\sigma^2=1, \beta=5.125$) but set the time scaling parameter $S = 5 * 10^{-6}$ to reflect the large range of chunk indices from 1 to 500.

\subsection{Time HDM Vectors}
Now that we have a continuous time representation, we need to somehow bind the representation with HDM's memory vectors. HDM itself uses the holographic reduced representations (HRR) framework \cite{plate_holographic_1995}. Plate shows how memory traces can be constructed by binding together two ``holographic" vectors $A, B \in \mathbb{R}^n$ using circular convolution ($\circledast$). From a memory trace $C = A \circledast B + ...$ and a cue $A$, a noisy version of $B$ can be recovered. Simply using the time vector itself as a holographic vector would violate HRR's assumption that bound vectors  are drawn from the same distribution. HDM's holographic vectors are simply drawn from the normal distribution, while the time vectors are oscillating outputs of $\mathbf{T}(t)$. Instead, we needed to treat the time vector as a kind of value being stored. The problem is that HDM, like ACT-R, works with discrete memory values even if it encodes them with continuous values. To encode the continuous time vector, 
\newcommand{\Tj}{\mathbf{\widetilde{T}}}
we adopted the fractional binding operation from \citeA{komer_neural_2019} and composed it with $\circledast$ from Plate to arrive at a holographic representation $\Tj$ of the time vector $\mathbf{T}$. 

\begin{align}
    \Tj = \mathcal{F}^{-1} \left\{ \sum_{l=1}^{320} \mathcal{F}(e_{t_l})^{\mathbf{T}(t)}\right\} \label{eq:Tj}
\end{align}

The function $\mathcal{F}$ above is the Fourier transform. The HRR vectors $e_{t_1} ... e_{t_{320}}$ are generated from the same distribution as HDM's environment holographic vectors. Next, we associate the chunk's time vector with each bit of memory in the chunk. We store these associations in $mt$ vectors -- Each HDM memory vector $m$ has an associated time-memory vector $mt$. This does double the memory usage but keeps the same space complexity class. When a chunk $s_1:v_1, ... s_n:v_n$ is added to memory, a $\Tj$ is created for it as per Equation \ref{eq:Tj} . The $mt$s are updated with the following rule:  
\begin{align}
    mt \longleftarrow mt + \Tj \circledast \sum_{c=1}^{n} s_c \circledast v_c
\end{align}
where $s_c$ and $v_c$ are corresponding slot-value pairs in a chunk.

This update rule, unlike the update rule for HDM's memory vectors, does not add noise. From an implementation standpoint, adding noise would interfere with the recalling of added traces in \citeA{plate_holographic_1995}, which already has some noise in reconstruction. From a cognitive processes representation standpoint, the noise to represent cognitively plausible time-memory errors is already represented in the time vectors themselves as discussed earlier.

\newcommand{\That}{\mathbf{\hat{T}}}
To retrieve the chunk $\chunk$ from cue $\cue = \{s'_1{:}v'_1, s'_2{:}v'_2, ... s'_m{:}v'_m\}$, we construct HRR $Q$ from $\cue$, take its inverse, and bind it with each memory-time vector $mt$ to obtain a noisy reconstruction $\That$ of the time HRR:
\begin{align}
Q = \sum_{c=1}^{m} s'_c \circledast v'_c \\
\That = mt \circledast Q^{-1}
\end{align}
The retrieval works because $\circledast$ is distributive, so $mt$ contains $\Tj \circledast s_c \circledast v_c$ for each slot:value pair. Note that the inverse operation used here is the approximate inverse from Section VII, Subsection C in \cite{plate_holographic_1995}.

We are left with $\That_1 ... \That_N$, the collection of reconstructed time vectors for each of the $N$ $mt$s. As per \citeA{brown_oscillator-based_2000}, we step through the time vector function $\mathbf{T}(t)$ from $t=1$ to the current time step in the model, comparing at each time step which of the reconstructed time vectors $\That$ have the greatest similarity with the current $\mathbf{T}(t)$. Depending on the method chosen --- taking the top $p$ results, all results above a threshold, or results meeting the previous two criteria --- we end up with a list of possible slots and values making up a chunk. HDM tracks which memory vectors are slots, so we keep the $mt$s corresponding to slots and then build a complete cue $\mathbf{q} = \{ s_1:? ... s_n:?\}$ that contains every slot. The chunk can then be requested using multiple unknown request chaining, mentioned at the beginning of this section, with HDM's built-in retrieval mechanism. Thus, we are in principle able to retrieve the entire chunk.

It's worth noting that, like HDM, the time complexity scales with the number of unique values rather than the total number of chunks. Still, an option to only turn on slot pulling encoding for chunks of interest so only a subset of relevant time-memory vectors are stored would reduce the retrieval times. However, remembering chunks (e.g. sentences) stored during the corpus adding step could be useful for some modeller applications as well.

\section{Preliminary Results and Discussion}
\subsection{Time Encodings}
\begin{figure}
    \centering
    \includegraphics[width=1\linewidth]{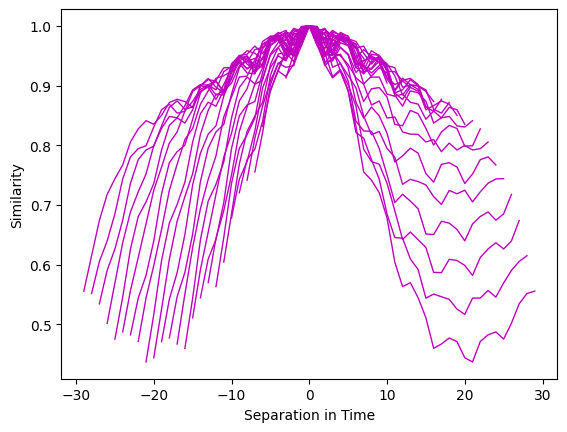}
    
    \caption{The dot product similarity between our time vector $\mathbf{T}(t') \boldsymbol{\cdot} \mathbf{T}(t'') $ (y axis)  and $t' - t''$ on the x axis. It peaks at 1 (comparing identical vectors) and has smaller local maxima  to represent noise in serial memory recall.} 
    \label{fig:better-temp-sim}
\end{figure}
\begin{figure}
    \centering
    \includegraphics[width=1\linewidth]{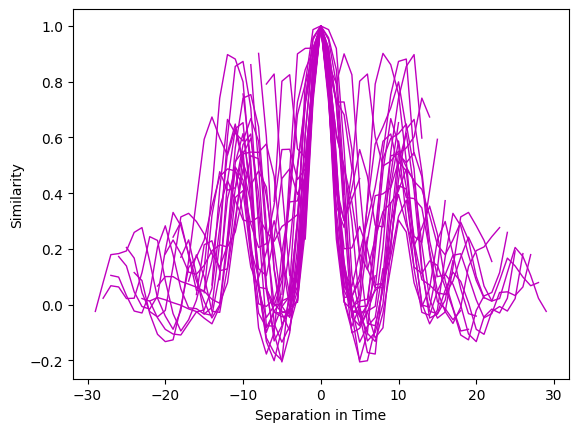}
    
    \caption{A self-similarity graph like Fig. \ref{fig:better-temp-sim} but with much larger ``bumps," indicating a noisier representation of serial memory.}
    \label{fig:worse-temp-sim}
\end{figure}
\begin{figure}
    \centering
    \includegraphics[width=1\linewidth]{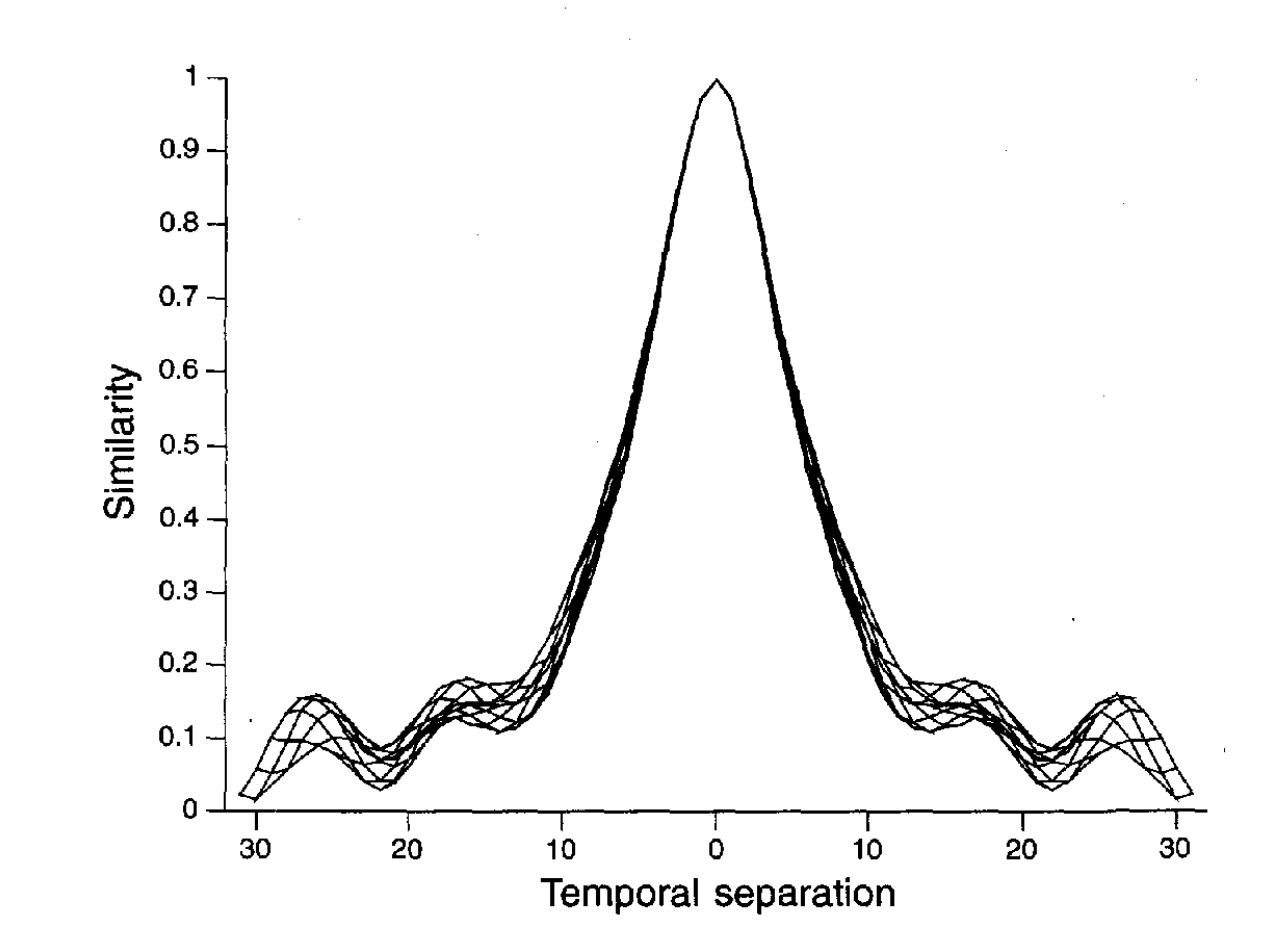}
    \caption{The (averaged) self-similarity of time vectors in \protect\citeA{brown_oscillator-based_2000}. Plot reproduced from Figure 7B of \protect\cite{brown_oscillator-based_2000}}
    \label{fig:brown-similarity}
\end{figure}

We generated two time vector functions using exactly the same set of parameters: $\sigma^2 = 1$ for Equation \ref{brown-normal}, $S=10^{-5}$ for Equation \ref{brown-theta}, and $\beta=5.125$ as the exponential distribution scaling parameter to sample four oscillators from the 15 oscillator functions to generate each element in Equation \ref{oscillator-prod}. The time vector has 320 elements as per our adaptation of \citeA{brown_oscillator-based_2000}. 

In order to evaluate the quality of the time vector representations, we plotted a similarity function $\mathbf{T}(t') \boldsymbol{\cdot} \mathbf{T}(t'')$ for all $t', t'' \in \{1, 2, .. 30\}$. The similarity function $\boldsymbol{\cdot}$ is the dot product, which is equivalent to cosine similarity here since the vectors are normalized. The similarity plots are shown in Figure \ref{fig:better-temp-sim} and Figure \ref{fig:worse-temp-sim}.

As noted earlier, a similarity function between time vectors taken at subsequent time steps should show two properties: (1) a gradual decrease as the times grow further apart and (2) some oscillation to represent the cognitive error that arises from time's hierarchical representations. Our time encodings do show both of these properties, but the ``bumps" that represent the error are more pronounced than in self-similarity plot of \citeA{brown_oscillator-based_2000}, reproduced here in Figure \ref{fig:brown-similarity}. Both plots are over the same time step range and equal corresponding parameters as our plot. A possible explanation is that the dot product between two large vectors rather than the average dot product between two sets of smaller vectors results in more noise, which \citeA{brown_oscillator-based_2000} mentioned as a justification for his choice of the latter representation. 

Furthermore, one can observe a stark difference between the self-similarity of Figure \ref{fig:better-temp-sim} and Figure \ref{fig:worse-temp-sim} despite having oscillator parameters generated from the same probability distributions with the same parameters. In our tests, neither increasing nor decreasing $\sigma^2$ had much consistent effect on how much variability appeared between the self-similarity plots of time vector functions. Increasing or decreasing the exponential distribution scale parameter $\beta$ to respectively favor or disfavor slower oscillators had surprisingly little effect either on the variance between time functions or on the level of noise any particular time function showed in self-similarity plots. Further study is needed to quantify the self-similarity noise in a single comparable metric and decide which parameters yield representations that best fit observed cognitive effects in serial order.

\section{Conclusion}
So far, we have described how full-chunk recall can be supported without needing to store the chunks themselves nor a data structure scaling in size with the number of associations between values in the chunk. The time vector representations we have implemented so far are noisy but still provide distinct representations for distinct chunks. Work to measure slot pulling effectiveness across a wider variety of models is ongoing.

In the future, we want to try another time encoding for associating chunks together besides neural oscillators. We also want to examine different chunk retrieval mechanisms. So far, we take the two inputs —reconstructed time vectors per time-memory vector and the time vectors themselves — and compute the similarity between them as the only ranking criterion. However, we should also expect that the time vector reconstructions associated with the same chunk ought to cluster together in vector space. Perhaps the true time vectors represent cluster centers and a clustering measure could aid accurate and precise retrieval. 

The vector-symbolic approach used here also presents an opportunity for more straightforward considerations of ways to integrate generative models within ACT-R. Such an integration would hold implications for expanded sociocultural representations in a cognitive architecture like ACT-R, as argued by \citeA{dancy_cogarch-genai_2023}. As previously mentioned, having these sociocultural representations in memory would also give an opportunity to explore the impacts of sociocultural structures on decision making behavior given the interdependence between memory and decisions. We plan to explore interactions between sociocultural structures, memory, and decision making using HDM (and eventually generative models) in the context of IBL.  

Adequately representing these sociocultural structures in memory requires memory representations to be scaled up efficiently. At the same time, there is value in making production systems work reliably with vector memory; production rules give explainable reasoning for decision making that generative models often struggle with. Full-chunk recall is needed to support flexible, reliable production systems interacting with vector-based memory. Also important is that full-chunk recall makes sure memories specific to an agent rather than its broader worldview are used at the appropriate times during decision making. The full-chunk recall mechanism and text-to-memory pipeline explained here hold promise for computational cognitive architectures like ACT-R to model the cognitive role that social structures play in human decision making.

\section{Acknowledgments}
 This material is based upon work supported by the AI Research Institutes Program funded by the National Science Foundation under the AI Institute for Societal Decision Making (NSF AI-SDM), Award No. 2229881. The graphs were generated with matplotlib \cite{matplotlib_2007}.



\bibliographystyle{apacite}

\setlength{\bibleftmargin}{.125in}
\setlength{\bibindent}{-\bibleftmargin}

\bibliography{refs}

\begin{thebibliography}{}

\bibitem [\protect \citeauthoryear {%
{ACT-R Research Group}%
}{%
{ACT-R Research Group}%
}{%
{\protect \APACyear {2023}}%
}]{%
act-r_research_group_act-r_2023}
\APACinsertmetastar {%
act-r_research_group_act-r_2023}%
\begin{APACrefauthors}%
{ACT-R Research Group}.%
\end{APACrefauthors}%
\unskip\
\newblock
\APACrefYearMonthDay{2023}{{\APACmonth{07}}}{}.
\newblock
\APACrefbtitle {{ACT}-{R} 7.27.9 software.} {{ACT}-{R} 7.27.9 software.}
\newblock
\begin{APACrefURL} [{2023-12-06}]\url{https://act-r.psy.cmu.edu/act-r-7-27-9-software/} \end{APACrefURL}
\PrintBackRefs{\CurrentBib}

\bibitem [\protect \citeauthoryear {%
Brown%
, Preece%
\BCBL {}\ \BBA {} Hulme%
}{%
Brown%
\ \protect \BOthers {.}}{%
{\protect \APACyear {2000}}%
}]{%
brown_oscillator-based_2000}
\APACinsertmetastar {%
brown_oscillator-based_2000}%
\begin{APACrefauthors}%
Brown, G\BPBI D\BPBI A.%
, Preece, T.%
\BCBL {}\ \BBA {} Hulme, C.%
\end{APACrefauthors}%
\unskip\
\newblock
\APACrefYearMonthDay{2000}{}{}.
\newblock
{\BBOQ}\APACrefatitle {Oscillator-{Based} {Memory} for {Serial} {Order}} {Oscillator-{Based} {Memory} for {Serial} {Order}}.{\BBCQ}
\newblock
\APACjournalVolNumPages{Psychological Review}{107}{1}{127--181}.
\PrintBackRefs{\CurrentBib}

\bibitem [\protect \citeauthoryear {%
Dancy%
\ \BBA {} Workman%
}{%
Dancy%
\ \BBA {} Workman%
}{%
{\protect \APACyear {2023}}%
}]{%
dancy_cogarch-genai_2023}
\APACinsertmetastar {%
dancy_cogarch-genai_2023}%
\begin{APACrefauthors}%
Dancy, C\BPBI L.%
\BCBT {}\ \BBA {} Workman, D.%
\end{APACrefauthors}%
\unskip\
\newblock
\APACrefYearMonthDay{2023}{}{}.
\newblock
{\BBOQ}\APACrefatitle {On integrating generative models into cognitive architectures for improved computational sociocultural representations} {On integrating generative models into cognitive architectures for improved computational sociocultural representations}{\BBCQ}\ [Conference Proceedings].
\newblock
\BIn{} \APACrefbtitle {AAAI Fall Symposium Series} {Aaai fall symposium series}\ (\BVOL~2, \BPG~256-261).
\newblock
\APACaddressPublisher{Washington, DC}{AAAI Press}.
\PrintBackRefs{\CurrentBib}

\bibitem [\protect \citeauthoryear {%
Gonzalez%
, Lerch%
\BCBL {}\ \BBA {} Lebiere%
}{%
Gonzalez%
\ \protect \BOthers {.}}{%
{\protect \APACyear {2003}}%
}]{%
gonzalez_ibl_2003}
\APACinsertmetastar {%
gonzalez_ibl_2003}%
\begin{APACrefauthors}%
Gonzalez, C.%
, Lerch, J\BPBI F.%
\BCBL {}\ \BBA {} Lebiere, C.%
\end{APACrefauthors}%
\unskip\
\newblock
\APACrefYearMonthDay{2003}{}{}.
\newblock
{\BBOQ}\APACrefatitle {Instance-based learning in dynamic decision making} {Instance-based learning in dynamic decision making}{\BBCQ}\ [Journal Article].
\newblock
\APACjournalVolNumPages{Cognitive Science}{27}{4}{591-635}.
\newblock
\begin{APACrefDOI} \doi{10.1207/s15516709cog2704_2} \end{APACrefDOI}
\PrintBackRefs{\CurrentBib}

\bibitem [\protect \citeauthoryear {%
Hunter%
}{%
Hunter%
}{%
{\protect \APACyear {2007}}%
}]{%
matplotlib_2007}
\APACinsertmetastar {%
matplotlib_2007}%
\begin{APACrefauthors}%
Hunter, J\BPBI D.%
\end{APACrefauthors}%
\unskip\
\newblock
\APACrefYearMonthDay{2007}{{\APACmonth{05}}}{}.
\newblock
{\BBOQ}\APACrefatitle {Matplotlib: {A} {2D} {Graphics} {Environment}} {Matplotlib: {A} {2D} {Graphics} {Environment}}.{\BBCQ}
\newblock
\APACjournalVolNumPages{Computing in Science \& Engineering}{9}{3}{90--95}.
\newblock
\APACrefnote{Conference Name: Computing in Science \& Engineering}
\newblock
\begin{APACrefDOI} \doi{10.1109/MCSE.2007.55} \end{APACrefDOI}
\PrintBackRefs{\CurrentBib}

\bibitem [\protect \citeauthoryear {%
Kelly%
}{%
Kelly%
}{%
{\protect \APACyear {2020}}%
}]{%
kelly_ecphoryhdm_2020}
\APACinsertmetastar {%
kelly_ecphoryhdm_2020}%
\begin{APACrefauthors}%
Kelly, M\BPBI A.%
\end{APACrefauthors}%
\unskip\
\newblock
\APACrefYearMonthDay{2020}{{\APACmonth{07}}}{}.
\newblock
\APACrefbtitle {ecphory/{HDM}.} {ecphory/{HDM}.}
\newblock
\begin{APACrefURL} [{2023-09-25}]\url{https://github.com/ecphory/HDM} \end{APACrefURL}
\newblock
\APACrefnote{original-date: 2020-03-29T22:27:50Z}
\PrintBackRefs{\CurrentBib}

\bibitem [\protect \citeauthoryear {%
Kelly%
, Arora%
, West%
\BCBL {}\ \BBA {} Reitter%
}{%
Kelly%
\ \protect \BOthers {.}}{%
{\protect \APACyear {2020}}%
}]{%
kelly_holographic_2020}
\APACinsertmetastar {%
kelly_holographic_2020}%
\begin{APACrefauthors}%
Kelly, M\BPBI A.%
, Arora, N.%
, West, R\BPBI L.%
\BCBL {}\ \BBA {} Reitter, D.%
\end{APACrefauthors}%
\unskip\
\newblock
\APACrefYearMonthDay{2020}{}{}.
\newblock
{\BBOQ}\APACrefatitle {Holographic {Declarative} {Memory}: {Distributional} {Semantics} as the {Architecture} of {Memory}} {Holographic {Declarative} {Memory}: {Distributional} {Semantics} as the {Architecture} of {Memory}}.{\BBCQ}
\newblock
\APACjournalVolNumPages{Cognitive Science}{44}{11}{e12904}.
\newblock
\begin{APACrefDOI} \doi{10.1111/cogs.12904} \end{APACrefDOI}
\PrintBackRefs{\CurrentBib}

\bibitem [\protect \citeauthoryear {%
Komer%
, Stewart%
, Voelker%
\BCBL {}\ \BBA {} Eliasmith%
}{%
Komer%
\ \protect \BOthers {.}}{%
{\protect \APACyear {2019}}%
}]{%
komer_neural_2019}
\APACinsertmetastar {%
komer_neural_2019}%
\begin{APACrefauthors}%
Komer, B.%
, Stewart, T\BPBI C.%
, Voelker, A\BPBI R.%
\BCBL {}\ \BBA {} Eliasmith, C.%
\end{APACrefauthors}%
\unskip\
\newblock
\APACrefYearMonthDay{2019}{{\APACmonth{07}}}{}.
\newblock
{\BBOQ}\APACrefatitle {A neural representation of continuous space using fractional binding} {A neural representation of continuous space using fractional binding}.{\BBCQ}
\newblock
\APACaddressPublisher{Montreal, Canada}{}.
\PrintBackRefs{\CurrentBib}

\bibitem [\protect \citeauthoryear {%
Lukosch%
, Bekebrede%
, Kurapati%
\BCBL {}\ \BBA {} Lukosch%
}{%
Lukosch%
\ \protect \BOthers {.}}{%
{\protect \APACyear {2018}}%
}]{%
lukosch_scientific_2018}
\APACinsertmetastar {%
lukosch_scientific_2018}%
\begin{APACrefauthors}%
Lukosch, H\BPBI K.%
, Bekebrede, G.%
, Kurapati, S.%
\BCBL {}\ \BBA {} Lukosch, S\BPBI G.%
\end{APACrefauthors}%
\unskip\
\newblock
\APACrefYearMonthDay{2018}{{\APACmonth{06}}}{}.
\newblock
{\BBOQ}\APACrefatitle {A {Scientific} {Foundation} of {Simulation} {Games} for the {Analysis} and {Design} of {Complex} {Systems}} {A {Scientific} {Foundation} of {Simulation} {Games} for the {Analysis} and {Design} of {Complex} {Systems}}.{\BBCQ}
\newblock
\APACjournalVolNumPages{Simulation \& Gaming}{49}{3}{279--314}.
\newblock
\begin{APACrefDOI} \doi{10.1177/1046878118768858} \end{APACrefDOI}
\PrintBackRefs{\CurrentBib}

\bibitem [\protect \citeauthoryear {%
Newell%
}{%
Newell%
}{%
{\protect \APACyear {1990}}%
}]{%
newell_utc_1990}
\APACinsertmetastar {%
newell_utc_1990}%
\begin{APACrefauthors}%
Newell, A.%
\end{APACrefauthors}%
\unskip\
\newblock
\APACrefYear{1990}.
\newblock
\APACrefbtitle {Unified theories of cognition} {Unified theories of cognition}\ [Book].
\newblock
\APACaddressPublisher{Cambridge, MA, USA}{Harvard University Press}.
\PrintBackRefs{\CurrentBib}

\bibitem [\protect \citeauthoryear {%
Plate%
}{%
Plate%
}{%
{\protect \APACyear {1995}}%
}]{%
plate_holographic_1995}
\APACinsertmetastar {%
plate_holographic_1995}%
\begin{APACrefauthors}%
Plate, T.%
\end{APACrefauthors}%
\unskip\
\newblock
\APACrefYearMonthDay{1995}{{\APACmonth{05}}}{}.
\newblock
{\BBOQ}\APACrefatitle {Holographic reduced representations} {Holographic reduced representations}.{\BBCQ}
\newblock
\APACjournalVolNumPages{IEEE Transactions on Neural Networks}{6}{3}{623--641}.
\newblock
\begin{APACrefDOI} \doi{10.1109/72.377968} \end{APACrefDOI}
\PrintBackRefs{\CurrentBib}

\bibitem [\protect \citeauthoryear {%
Prather%
\ \protect \BOthers {.}}{%
Prather%
\ \protect \BOthers {.}}{%
{\protect \APACyear {2022}}%
}]{%
prather_what_2022}
\APACinsertmetastar {%
prather_what_2022}%
\begin{APACrefauthors}%
Prather, R\BPBI W.%
, Benitez, V\BPBI L.%
, Brooks, L\BPBI K.%
, Dancy, C\BPBI L.%
, Dilworth-Bart, J.%
, Dutra, N\BPBI B.%
\BDBL {}Thomas, A\BPBI K.%
\end{APACrefauthors}%
\unskip\
\newblock
\APACrefYearMonthDay{2022}{}{}.
\newblock
{\BBOQ}\APACrefatitle {What {Can} {Cognitive} {Science} {Do} for {People}?} {What {Can} {Cognitive} {Science} {Do} for {People}?}{\BBCQ}
\newblock
\APACjournalVolNumPages{Cognitive Science}{46}{6}{e13167}.
\newblock
\begin{APACrefDOI} \doi{10.1111/cogs.13167} \end{APACrefDOI}
\PrintBackRefs{\CurrentBib}

\bibitem [\protect \citeauthoryear {%
Ritter%
, Tehranchi%
\BCBL {}\ \BBA {} Oury%
}{%
Ritter%
\ \protect \BOthers {.}}{%
{\protect \APACyear {2019}}%
}]{%
ritter_act-r_2019}
\APACinsertmetastar {%
ritter_act-r_2019}%
\begin{APACrefauthors}%
Ritter, F\BPBI E.%
, Tehranchi, F.%
\BCBL {}\ \BBA {} Oury, J\BPBI D.%
\end{APACrefauthors}%
\unskip\
\newblock
\APACrefYearMonthDay{2019}{}{}.
\newblock
{\BBOQ}\APACrefatitle {{ACT}-{R}: {A} cognitive architecture for modeling cognition} {{ACT}-{R}: {A} cognitive architecture for modeling cognition}.{\BBCQ}
\newblock
\APACjournalVolNumPages{WIREs Cognitive Science}{10}{3}{e1488}.
\newblock
\begin{APACrefDOI} \doi{10.1002/wcs.1488} \end{APACrefDOI}
\PrintBackRefs{\CurrentBib}

\bibitem [\protect \citeauthoryear {%
Sarica%
\ \BBA {} Luo%
}{%
Sarica%
\ \BBA {} Luo%
}{%
{\protect \APACyear {2021}}%
}]{%
sarica_stopwords_2021}
\APACinsertmetastar {%
sarica_stopwords_2021}%
\begin{APACrefauthors}%
Sarica, S.%
\BCBT {}\ \BBA {} Luo, J.%
\end{APACrefauthors}%
\unskip\
\newblock
\APACrefYearMonthDay{2021}{{\APACmonth{08}}}{}.
\newblock
{\BBOQ}\APACrefatitle {Stopwords in technical language processing} {Stopwords in technical language processing}.{\BBCQ}
\newblock
\APACjournalVolNumPages{PLOS ONE}{16}{8}{e0254937}.
\newblock
\APACrefnote{Publisher: Public Library of Science}
\newblock
\begin{APACrefDOI} \doi{10.1371/journal.pone.0254937} \end{APACrefDOI}
\PrintBackRefs{\CurrentBib}

\end{thebibliography}

\end{document}